\documentclass[conference]{IEEEtran}
\usepackage{spconf,amsmath,graphicx}

\let\labelindent\relax
\usepackage{enumitem}
\setlist{nosep, leftmargin=14pt}

\usepackage{mwe} 

\title{Interactive segmentation using U-Net with weight map and dynamic user interactions}
\name{Ragavie Pirabaharan and Naimul Khan} 
\address{Ryerson University, Toronto, ON}
\begin{document}
\abovedisplayskip=0.3cm
\abovedisplayshortskip=-0.1cm
\belowdisplayskip=0.5cm
\belowdisplayshortskip=0.2cm

\maketitle

\begin{abstract}
Interactive segmentation has recently attracted attention for specialized tasks where expert input is required to further enhance the segmentation performance. In this work, we propose a novel interactive segmentation framework, where user clicks are dynamically adapted in size based on the current segmentation mask. The clicked regions form a weight map and are fed to a deep neural network as a novel weighted loss function. 
To evaluate our loss function, an interactive U-Net (IU-Net) model which applies both foreground and background user clicks as the main method of interaction is employed. We train and validate on the BCV dataset, while testing on spleen and colon cancer CT images from the MSD dataset to improve the overall segmentation accuracy in comparison to the standard U-Net using our weighted loss function. Applying dynamic user click sizes increases the overall accuracy by 5.60\% and 10.39\% respectively by utilizing only a single user interaction. 
\end{abstract}
%
%
\section{Introduction}
\label{sec:intro}

Recently, interactive image segmentation has used deep learning with Convolutional Neural Networks (CNNs) to learn high level semantics from images \cite{Hesamian}. 
Current innovations in interactive medical image segmentation that use CNNs include but are not limited to DeepIGeoS \cite{Wang_2019}, Sakinis et al. \cite{sakinis} and BIFSeg \cite{Wang_2018}. DeepIGeoS \cite{Wang_2019}, uses CNNs and provides geodesic distance transforms of user provided scribbles to the CNN as a form of user interaction. The performance of this model can depreciate for unseen objects and works only with objects that were previously seen. Sakinis et al. \cite{sakinis} takes user input through the form of foreground and background clicks, combined with an image as an input into an U-Net model. We will call this approach Interactive U-Net or IU-Net. This model is the most user-friendly approach by employing the simplest form of user interaction through clicks only, and hence serves as the baseline to our proposed approach.  BIFSeg \cite{Wang_2018} uses bounding boxes as user input and image-specific fine tuning through user provided scribbles to segment unseen objects in different MR sequences. This method is considered time consuming since it requires additional user interaction aside from the initial bounding boxes applied to the input image through the form of fine tuning for each test image to improve segmentation accuracy.

In cases, where there are large class imbalances in the data set, weights and fine tuning are implemented into the loss function. The popular U-Net \cite{UNet} uses weighted loss function with weight maps. 
Low weights were assigned to the center of the region as well as the background, while pixels nearing the boundary of the region contained higher weights  \cite{UNet}\cite{Cells}. This alleviates the imbalance within the classes of the dataset, while allowing the model to capture the borders of the ROI \cite{Cells}. 
BIFSeg \cite{Wang_2018} employed the use of weighted loss functions in an interactive segmentation framework. 
Users provided extra scribbles on under segmented locations, which were given high pixel weights and factored into the loss function as a weight map during only testing. However, the additional tuning was only applied within the testing stage and the pixel size of the scribbles were fixed. All existing works use a fixed click size, while the ROI might vary in size greatly across datasets. Therefore, to alleviate this, we introduce an adaptive click size which is adjusted automatically based on the size of the current ROI.

The main contributions of this work are:

\begin{enumerate}
  \item [1.]We produce a weighted loss function created by concatenating both weighted foreground clicks provided by the user with weighted gaussian pixel maps on the image.
  \item [2.] We produce the first known introduction of dynamically changing click sizes in accordance with the size of the region during training to help with intra-class variation.
\end{enumerate}

The model used for testing consists of U-Net and IU-Net. Experiments were trained and validated on the BCV \footnote{https://www.synapse.org/Synapse:syn3193805/wiki/217789} dataset held at MICCAI 2015 for the ``multi-atlas labeling: beyond the cranial vault'' challenge. Testing was completed on the spleen and colon cancer CT data from the MSD \footnote{http://medicaldecathlon.com/} dataset \cite{MSD} held at MICCAI 2018. 
With just a single user click combined with our dynamic click size and weighted loss function, we achieve 5.60\% and 10.39\% increase in DSC for the two datasets, respectively.

\section{Method}
Motivated by the findings of \cite{sakinis}, we decided to employ their interactive segmentation framework, which uses a base U-Net model and takes user input through mouse clicks for foreground and background segmentation. The 2D segmentation network was trained to interactively segment an ROI at a time.  The model is trained with concatenated 2D images consisting of raw 2D image and network guidance images comprised of user guided foreground and background clicks for interactive segmentation input, as seen in Figure 1. We use a weighted loss function comprised of weight maps and clicks for both training and testing. 

\begin{figure}[htp]
    \centering
    \includegraphics[width=8.5 cm]{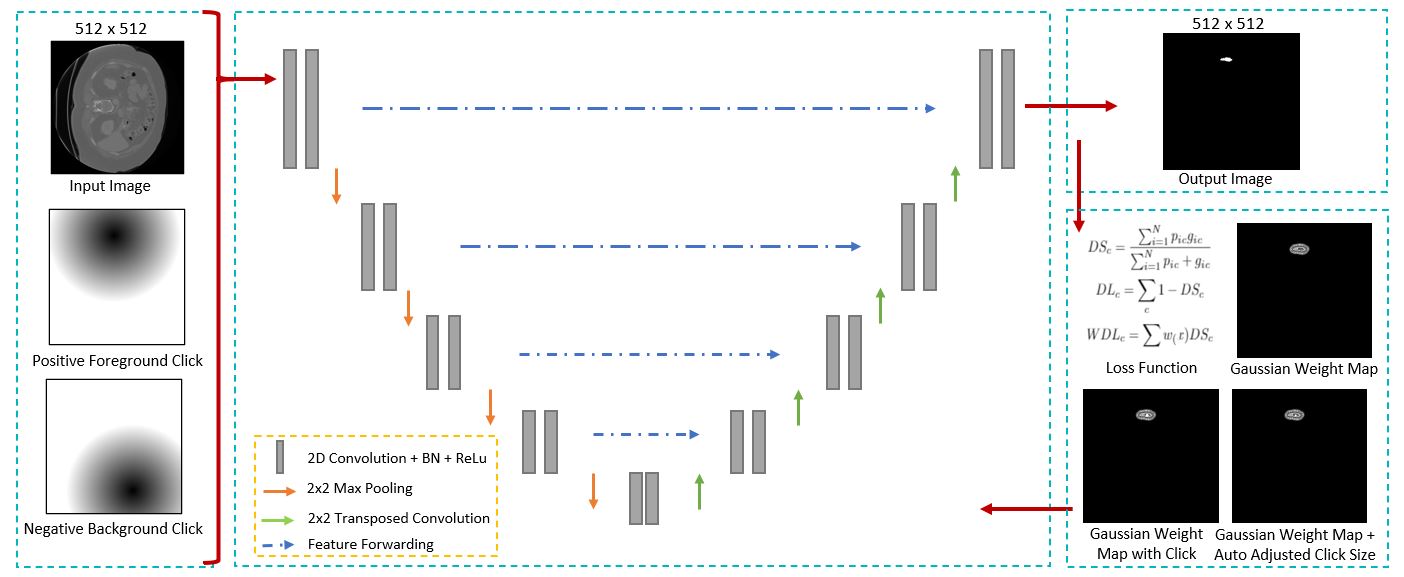}
    \caption{Model schematic including IU-Net, click input images and weighted loss function}
    \label{fig:galaxy}
\end{figure}

User guidance is provided through the form of simulated clicks on the image to guide the network to segment an ROI. Depending on the location of the clicks, the area selected is determined to be a foreground or background. Foreground clicks are used to guide the network to a ROI or inform it to segment that area \cite{sakinis}. Background clicks are used to inform the system about over segmented and unnecessary locations \cite{sakinis}. Similar to \cite{sakinis}, simulated clicks were used on the images. Based upon the location of the ground truth and surrounding regions on the image, randomly placed foreground and background clicks were created. This method differs from the interaction method used in \cite{sakinis}, as they employed disparity maps for clicks. We believe randomized click event captures user behavior more accurately and accounts for inadvertent user errors that can happen during clicking on a small region. The locations of the clicks were saved as both a foreground and background distance map image as shown in Figure 1. Both the foreground and background distance map images are the same spatial size as the input image but have a pixel value wherever the clicks were simulated \cite{sakinis}. One user interaction, which consists of a positive click and negative click, and an original image are concatenated before being fed into the model.

We adopted the popular U-Net architecture \cite{UNet} as our baseline, which closely resembles the architecture in \cite{sakinis}. The IU-Net model is similar to U-Net and consists of 4 blocks for both up and down sampling as well as a bottleneck block. The encoder layer has each block containing two 2D convolutions with 3x3 kernels, ReLu activations, dropout layers, along with a max-pooling operation. The decoder layer has each block containing 2D transpose convolutions with a 3x3 kernel, two 2D convolutions with 3x3 kernels and ReLu activation. At the final layer, there is a 1x1 convolution used to map out the final feature vectors of the image.

To assess segmentation maps, the dice score coefficient (DSC) is used by the medical community. As it is used as an overlap index, the 2-class dice score is widely used to compare an annotated ground truth map with a prediction map. The dice loss function is categorized as the minimization of the similarity between the ground truth and prediction map of the image \cite{abraham}. The dice loss function was empirically selected to be used for the following experiments in this paper, as it provided the best DSC when used with the model in comparison to other state of the art loss functions that were tested.

\begin{equation}
DL_c=\sum_{c}(1-\frac{\sum_{i=1}^{N} p_{ic} g_{ic}}{\sum_{i=1}^{N} p_{ic} + g_{ic}})
\end{equation}

As depicted in Equation 1, dice loss consists of both \(p_{ic}\) and \(g_{ic}\) which represents the predicted and ground truth respectively for each class \(c\), pixel \(i\) and number of pixels within an image \(N\). Dice loss has shown that it is tough on outliers, and acceptable for class imbalance issues \cite{abraham}. 

\subsection{Gaussian Weight Map and Fine Tuning}
\label{ssec:subhead}
To improve upon the accuracy of the loss function, we propose the factoring of a gaussian weight map in conjunction with high weighted pixel clicks provided by the user. We used a gaussian weight map for the loss function that was precomputed for each training image. Like the method used by U-Net \cite{UNet}, pixels near the edges of the segmentation boundaries were given higher weights than those near the centre of the ROI. The pixel weight assigned increases gradually from a weight of 0 at the center of the map to the highest weight of 10 at the edge. Gaussian was used instead of other weight map methods as it places higher weight allocation at the edge of the ROI for better segmentation. 

The addition of fine tuning in weighted loss functions was introduced in \cite{Wang_2018}, where user defined scribbles are provided to the loss function in the testing stage. Inspired by this, we investigate the use of combining user defined foreground clicks that were first presented as input into the model with the weight map. Through this interaction, user guided clicks are saved and each click is provided to both the input of the model and to the loss function simultaneously to improve accuracy. Combining both the dice loss function with the weight maps and user defined clicks provides us with our proposed weighted loss function:

\begin{equation}
w(c) = \max_{c \in i} d(i,c) \cdot \sum_{n=1}^{P}w(i)
\end{equation}

\begin{equation}
wDL_c=\sum_{c}w(c)(1-\frac{\sum_{i=1}^{N} p_{ic} g_{ic}}{\sum_{i=1}^{N} p_{ic} + g_{ic}})
\end{equation}

The gaussian weight map, represented by the function \(d(i,c)\), and user click, represented by \(w(i)\), are concatenated and is factored with the loss function as \(w(c)\). This weighted dice loss function, \(wDL_c\), is used in both the training and testing stage unlike BIFSeg \cite{Wang_2018}, which uses it in the latter. BIFSeg only applies it during testing since their loss function requires additional scribble based interaction aside from the bounding box based interaction given to the input of the model. Our method only requires one form of user interaction that is used for both the loss function and input of the model per image. User provided clicks have a much higher certainty than that of the surrounding pixels of the image and thus the weight associated to those pixel values was set at the highest weight of 10. Two types of pixel weights were tested; one with all the pixel weights being of equal value (value of 10), and the other where the pixel weights were of a gaussian nature. The gaussian weight maps used with or without user interaction are depicted in Figure 1.  

\subsection{Pixel Size of Click Based Fine Tuning}
\label{ssec:subhead}
To further achieve better segmentation accuracy with the weighted loss function, we investigate the pixel size of user clicks and the affect on segmentation. In our baseline experiments, the click size is set firmly at 5 pixels. Since class imbalance and large intra-class variation is present within the dataset, the size of the user clicks will not correspond correctly to the size of the ROI in most cases. We used alpha, \(\alpha\), to better represent the click sizes on the ROI when intra-class variation is present. This is expressed in our click size equation as:

\begin{equation}
cs = \alpha (cms)
\end{equation}

where the number of pixels within the class mask size, \(cms\), factored with \(\alpha\) determines the final pixel click size, \(cs\), for each image. We experiment with \(\alpha\) and observe the best performance among \(\alpha= \frac{1}{500}\) and \(\alpha = \frac{1}{800}\). 

As true mask size is unknown during test time, the mask size is computed by determining the ROI during run time. After the segmentation is obtained, the mask size of the ROI is factored with the alpha determining the click size during test. Combining \(\alpha\) with the number of pixels present within a class provides the dynamically changing click size for each image. Values larger than \(\frac{1}{500}\), had user click sizes exceed the bounds of the regions; smaller values were thus considered for proportionate reasons. Experiments were also conducted on clicks with a firm pixel size of 2 and 10 to show the improvements that dynamically changing click size has on segmentation quality. 

\section{Experiments}
We train our models on the 30 training volumes from the BCV dataset which depict a total of 13 abdominal organs. We test our approach on two datasets, spleen and colon cancer CT images, from the MSD Dataset open sourced in \cite{MSD} to determine our models ability to generalize. The spleen dataset consisted of 61 volumes separated into 41 training and 20 testing sets, whereas the colon dataset consisted of 190 volumes separated into 126 training and 64 testing sets. Only the training volumes were used for experimentation purposes. The average image size is 512 x 512 pixels, where the images contained segmentation of the colon or the spleen.

To present a fair evaluation of the weighted loss function and the effect that user click size has on the segmentation accuracy, the datasets are not augmented and do not have transfer learning conducted on it. 

\begin{table*}[ht]
\caption{Performance on MSD Colon and Spleen dataset}\label{tab1}
\centering
\begin{tabular}{ccccc}
\hline
\centering
Model & Weight\centering & Click Size & Spleen DSC & Colon DSC\\
\hline
\centering
    U-Net \centering & - \centering & 5 px & 70.23 & 63.91 \\
    IU-Net \cite{sakinis} \centering & -\centering &  5 px &  70.97 & 65.71\\
    IU-Net \centering & (G) Weight Map \centering & 5 px & 71.26 & 67.63\\
    IU-Net \centering & (G) Weight Map + (G) FG Click \centering & 5 px & 71.89 & 69.62 \\
    IU-Net \centering & (G) Weight Map + (EW) FG Click \centering & 5 px & 73.53 & 72.35\\
    IU-Net \centering & (G) Weight Map + (EW) FG Click \centering &  2 px & 73.19 & 72.58 \\
    IU-Net \centering & (G) Weight Map + (EW) FG Click \centering  & 10 px & 73.41 & 73.02\\
    IU-Net \centering & (G) Weight Map + (EW) FG Click \centering & \(\alpha = \frac{1}{500}\) \centering & 75.05 \centering & 74.06 \\
    IU-Net \centering & (G) Weight Map + (EW) FG Click \centering & \(\alpha = \frac{1}{800}\) \centering & \textbf{75.83} \centering & \textbf{74.30}\\
\hline
\end{tabular}
\end{table*}

\begin{table}[h]
  \centering
  \begin{tabular}{lllllll}
    \hline
           &   CT  & & DSC & & \\
    \hline
    Model & Clicks & 1 & 2 & 5 & 10 & 15 \\
    \hline
    Ours & Spleen & \bf{75.83} & \bf{77.70} & \bf{78.27} & \bf{79.49} & \bf{81.96}\\
    IU-Net & Spleen & 70.97 & 71.63 & 73.07 & 77.14 & 79.32\\
    Ours & Colon & \bf{74.30} & \bf{76.96} & \bf{77.49} & \bf{78.25} & \bf{80.87}\\
    IU-Net & Colon & 65.71 & 66.18 & 69.65 & 72.22 & 74.43\\
\hline
  \end{tabular}
  \caption{Performance achieved using experiment 9 on the MSD spleen and colon dataset with various interactions in comparison to baseline IU-Net performance [3]}\label{tab2}
  \vspace*{-0.290 cm}
\end{table}

We perform ablation study with 8 variations within the IU-Net and weighted loss function while comparing to the baseline U-Net trained with dice loss:
\vspace*{0.1 cm}
\begin{enumerate}
  \item [1.] U-Net trained with dice loss function
  \item [2.]IU-Net trained with dice loss function
  \item [3.]IU-Net trained with gaussian weight map and weighted dice loss function
  \item [4.] Experiment 3. with gaussian weighted clicks
  \item [5.] Experiment 3. with equal weighted clicks
  \item [6.]Experiment 5. with click size of 2 pixels
  \item [7.]Experiment 5. with click size of 10 pixels
  \item [8.]Experiment 5. with pixel click ratio of \(\alpha = \frac{1}{500}\)
  \item [9.]Experiment 5. with pixel click ratio of \(\alpha = \frac{1}{800}\)
\end{enumerate}
\vspace*{0.1 cm}

Experiments 3 to 5 explore the use of gaussian weight maps combined with user guided clicks that either have an equal weight (EW) of 10 or a gaussian (G) weight distribution. These are factored into the loss function to create the weighted dice loss function. All the clicks, unless explicitly stated, have a pixel size of 5. Experiments 6 to 9 are 4 case variations regarding the effect that pixel click size have on the overall DSC. Static clicks of pixel size 2, 5 and 10 along with dynamically adjusting click sizes with an \(\alpha = \frac{1}{500}\) and \( \alpha = \frac{1}{800}\) are shown. These ablation test results are documented in Table 1. Experiments to determine the validity of the loss function in situations where there is more than one click are conducted in Table 2, while comparisons are made to the results we generated using [3]'s model and dataset. This test was performed using 1, 2, 5, 10 and 15 user clicks on experiment 9 from Table 1. A grid search method was used to fine-tune the hyper parameters with the spleen dataset trained for 100 epochs with a batch size of 8 and colon cancer dataset trained for 100 epochs with a batch size of 2. All experiments are programmed using the Keras framework with the TensorFlow backend. 
\section{Results}
Table 1 indicates that IU-Net models trained with the weighted loss function in experiment 9 show a DSC increase by 5.60\% and 10.39\% in comparison to the baseline U-Net for the spleen and colon dataset. When comparing the original U-Net with the IU-Net trained with the weight loss function in experiment 5, there is a 3.30\% and 8.44\% increase in the DSC for both spleen and colon. We observe that incorporating a dynamically changing click size in Table 1 improves the overall DSC as it is better adjusted to intra-class variation. With the baseline being a click size of 5 pixels for dynamically changing click size, an \(\alpha = \frac{1}{800}\) enhanced the overall DSC by 2.30\%  and 1.95\% for the spleen and colon dataset. Finally, the weighted loss function was tested with multiple user interactions on the spleen and colon dataset in Table 2. All previous experiments in Table 1 were done with 1 user interaction. Table 2 shows that multiple user interactions on an image increases the overall accuracy of segmentation by a maximum of 6.13\% for spleen and 6.57\% for colon. When compared to the results obtained by replicating the model in \cite{sakinis}, the results were higher for our method when the number of user interactions increased.Though Table 2 depicts that our model performs better with multiple clicks, we would like to acknowledge that the comparison of our model to \cite{sakinis} is not completely fair due to dataset and architectural differences.

\section{Conclusion}
In this work, we propose a weighted loss function which utilises gaussian weight maps with foreground user clicks to improve semantic segmentation. Our experiments demonstrate that providing user guided clicks to the input of the interactive U-Net model as pixel weights and dynamically adjusting size in combination with a weight map, delivers an increase in accuracy by 5.60\% and 10.39\% for the spleen and colon dataset. We also improved the accuracy of the model by using dynamically adjusting pixel click sizes based upon the size of the class being investigated in the image. By using \(\alpha = \frac{1}{800}\), the accuracy of the model increased to 75.83\% from 73.53\%. This helped to address the large intra-class variation and class imbalance present within the dataset. Finally, our model was tested with multiple user interactions with the result increasing to 78.27\% with 5 user clicks and 81.96\% with 15 user clicks in comparison to 75.83\% given by 1 user click for the spleen dataset. Our proposed model with use of the loss function outperforms both the baseline U-Net and baseline interactive U-Net DSC. Overall, this model produces high accuracy with one user interaction and is efficient as it does not require extra user interactions other than the ones provided by the expert into the input. Future steps would include enhancing the model to increase accuracy when generalizing on unseen objects. 




\bibliographystyle{IEEEtran}
\bibliography{refs}

\end{document}